# “AI Is Turning Too Human”: How Teenagers Experience and Negotiate AI in Everyday Life

Jianfeng Zhu

Department of Computer Science, Kent State University, Kent, OH 44224, USA

Correspondence: jzhu10@kent.edu

## Abstract

Generative AI is rapidly entering adolescents’ everyday lives during a critical period of cognitive, social and emotional development. Yet its adoption is outpacing evidence on how adolescents themselves experience, understand and negotiate its expanding role in their lives. We examined AI-related discourse on r/teenagers from January 2023 to July 2026 using validated keyword-based retrieval and a human-in-the-loop, LLM-assisted thematic analysis. AI-related discussion increased substantially over time, and 11,083 analytically coded posts revealed eight interconnected domains of experience. Everyday and social use was most prevalent (36.8%), while discourse increasingly shifted toward authenticity, personal control and safety, and future human roles. Across domains, adolescents questioned when AI should support or substitute for human thinking and creativity, how conversational AI changes relationships and perceptions of agency, what can still be considered authentic, who controls personal information and representation, and what opportunities and roles should remain human. These findings position adolescent AI use not simply as technology adoption, but as an emerging negotiation over AI’s place and boundaries in everyday life. Supporting this transition will require developmentally appropriate AI literacy, psychological and social support, and AI systems and policies that protect adolescents’ agency, privacy, relationships and opportunities for human development.

## 1. Introduction

### 1.1 AI in Adolescents' Everyday Lives

Artificial intelligence has rapidly become embedded in adolescents’ everyday lives [1]. Recent nationally representative data from the United States show that 64% of teenagers aged 13–17 use AI chatbots, including approximately three in ten who report daily use [2]. A 2026 Common Sense Media survey found that 86% of youth aged 9–17 had used or interacted with some form of AI and nearly one-quarter did so daily [3]. Adolescents report turning to AI for information seeking (57%), schoolwork (54%), entertainment (47%), content summarization, and image or video creation and editing [2].

Increasingly, these interactions extend beyond task-oriented use into social and emotional life. Sixteen percent of adolescents report using chatbots for casual conversation and 12% for emotional support or advice [2], while 72% have used AI companions at least once [3]. AI is therefore becoming not only more prevalent in adolescents' lives, but also more deeply integrated across domains that involve learning, creativity, social interaction and emotional experience [4].

### 1.2 Emerging Concerns and Governance of Adolescent AI Use

The growing role of AI in adolescents' lives has raised questions about both its developmental benefits and potential risks. In education, generative AI can provide explanation, feedback and individualized support, but these benefits depend on how it is used: the OECD distinguishes improved task performance from genuine learning gains and cautions that outsourcing cognitive work to AI may improve performance without strengthening learning [5]. Beyond learning, concerns extend to privacy, autonomy, safety, social interaction and psychological well-being. These issues increasingly appear in governance frameworks. UNESCO calls for human-centred, age-appropriate use of generative AI with explicit protections for privacy and human agency, while UNICEF's child-centred AI framework emphasizes safety, data and privacy, fairness, transparency, development and well-being, inclusion, and preparation for an AI-mediated future [6,7].

Yet governance is developing alongside a rapidly changing set of adolescent practices [8]. Existing national surveys have provided important evidence on adolescents' AI use across predefined domains, including schoolwork, trust and authenticity, companionship, emotional support and safety [1,4,5,7]. However, because these studies necessarily examine experiences through researcher-specified questions and categories, less is known about what concerns and meanings emerge when adolescents discuss AI in their own terms.

### 1.3 Naturally Occurring Peer Discourse

Naturally occurring online discourse offers a complementary perspective to survey-based evidence [9,10]. Whereas surveys capture responses to researcher-defined questions, spontaneous peer discussions can reveal which aspects of AI adolescents themselves raise, how they interpret their experiences, and where uncertainty, disagreement and emerging norms arise [11,12]. This distinction may be particularly valuable in adolescent AI research, where rapidly changing technologies can generate new practices, concerns and social norms before they are incorporated into established measures.

Social media therefore provides an opportunity to examine which aspects of AI adolescents themselves raise, how they describe their experiences, and where uncertainty, disagreement and emerging boundaries arise.

### 1.4 Present Study

The present study examines naturally occurring AI-related discourse on r/teenagers using an inductive, human-in-the-loop thematic approach. Rather than imposing predefined domains of AI use or categories of benefit and risk, we examine the patterns of meaning that emerge from adolescents' own accounts. We ask: **How do adolescents experience, interpret, and negotiate the role of AI in their everyday lives?**

## 2 Related Work

### 2.1 Human–AI interaction is rapidly evolving

The emergence of generative AI has expanded the scope of human–AI interaction. Large language models enable open-ended interaction through natural language, supporting conversational, informational and creative activities and reshaping established forms of human–AI interaction [13]. Research on generative-AI adoption further suggests that use is shaped by factors including perceived usefulness or performance expectancy, task fit, trust and habit [14–18].

This shift is also changing the social possibilities of human–AI interaction. Conversational AI can provide responsive and personalized interaction and, in some contexts, be experienced as a social rather than exclusively instrumental partner. Emerging evidence shows that interaction with conversational AI can elicit self-disclosure, perceived interpersonal closeness, social connection and emotional attachment, although these experiences vary across individuals and contexts [19–21]. AI adoption therefore involves more than the uptake of a new tool; it introduces forms of interaction whose roles and boundaries are still developing.

### 2.2 Research approaches to technology adoption and experience

Research on technology adoption has traditionally relied on researcher-elicited methods. Frameworks such as the Technology Acceptance Model (TAM) and Unified Theory of Acceptance and Use of Technology (UTAUT) examine how factors including perceived usefulness, ease of use, performance expectancy and social influence shape technology acceptance and use [22,23]. Surveys operationalize such constructs at scale, while interviews and focus groups provide richer accounts of users' motivations, interpretations, and experiences [24].

These approaches necessarily structure evidence around researcher-defined constructs or questions. Naturally occurring discourse provides a complementary perspective by capturing experiences and meanings expressed outside direct researcher elicitation [23]. This distinction may be particularly important for generative AI, where capabilities, uses and social practices are evolving rapidly and may precede their representation in established measures.

### 2.3 AI use during adolescence

Adolescence is a particularly important period in which to examine these changes [25]. It is marked by continued development of autonomy, identity, social relationships and higher-order cognitive capacities, alongside increasing sensitivity to peer evaluation and social belonging [15,26,27]. Digital technologies are already embedded within many of these developmental processes, shaping how adolescents communicate, seek information, express identity and maintain relationships [28]. Generative AI introduces a distinctive element into this environment because the technology can participate directly in activities that are themselves developmentally salient, including learning, creative expression, self-disclosure and social interaction [29].

Existing research has begun to document adolescents' use of AI across education, companionship, emotional support, privacy and safety, and AI-generated content [15,30–34]. However, these domains have largely been examined separately and through researcher-defined measures. Less is known about which experiences adolescents themselves raise when discussing AI spontaneously, how concerns across these domains intersect, and how they change as AI becomes embedded in everyday life. A naturalistic, inductive approach can complement existing evidence by examining adolescent AI experience without specifying these domains in advance.

## 3 Methods

### 3.1 Data Source and AI-Related Post Identification

We collected Reddit submissions from r/teenagers and used June 2025 as a development and validation month for constructing the AI-related corpus [35]. For each submission, the title and self-text were cleaned and concatenated into a single post-level text field. AI-related posts were identified using a keyword dictionary targeting explicit AI terminology, generative-AI systems, AI-chatbot and companion platforms, and AI-generated media. Ambiguous terms were excluded unless they occurred in an explicit AI context. The retrieval procedure was iteratively refined and validated against human relevance coding in the June 2025 development sample before being applied to the full study period. The finalized keyword approach achieved 91.5% precision, 85.5% recall and an F1 score of 88.4% against human coding. Full keyword definitions, development procedures and validation results are reported in Appendix 1.

### 3.2 Multi-agent, Human-in-the-Loop Thematic Analysis

We conducted a multi-agent, human-in-the-loop inductive thematic analysis to identify recurring patterns of meaning in adolescents' AI-related discourse without imposing a predefined thematic taxonomy (Fig. 1). This design treated LLMs as analytic assistants rather than autonomous qualitative analysts, consistent with emerging work showing that LLMs can support coding and theme development at scale while researcher interpretation and validation remain important for thematic coherence, contextual interpretation and methodological rigor [36].

Theme development was conducted on a year-stratified discovery sample of 600 AI-related posts (200 each from 2023, 2024 and 2025). All LLM agents used GPT-5.6-luna [37], accessed programmatically through the OpenAI API, with separate prompts defining distinct analytical roles. Three agents independently coded each post from complementary semantic/descriptive (Agent A), experiential/interpretive (Agent B), and relational/normative (Agent C) perspectives. Each generated inductive codes grounded in verbatim textual evidence without access to the other agents' outputs.

Codes from the three agents were consolidated and reviewed by a fourth agent (Agent D), which developed higher-order candidate themes across posts and analytical perspectives. A fifth agent (Agent E) critically evaluated these candidate themes for evidential support, coherence, overlap and scope, recommending retention, revision, merging, splitting or removal. The resulting thematic framework comprised eight themes derived through this multi-agent inductive process.

Researchers then reviewed the candidate themes and critical-review outputs against randomly sampled posts and representative posts associated with each candidate theme. This review assessed whether the proposed themes were interpretable, distinct, and adequately grounded in the underlying adolescent discourse. Based on this review, the research team confirmed and, where necessary, refined theme boundaries and labels, resulting in eight human-reviewed final themes. Full agent instructions, structured output specifications, and the thematic-development audit trail are provided in Appendix 2.

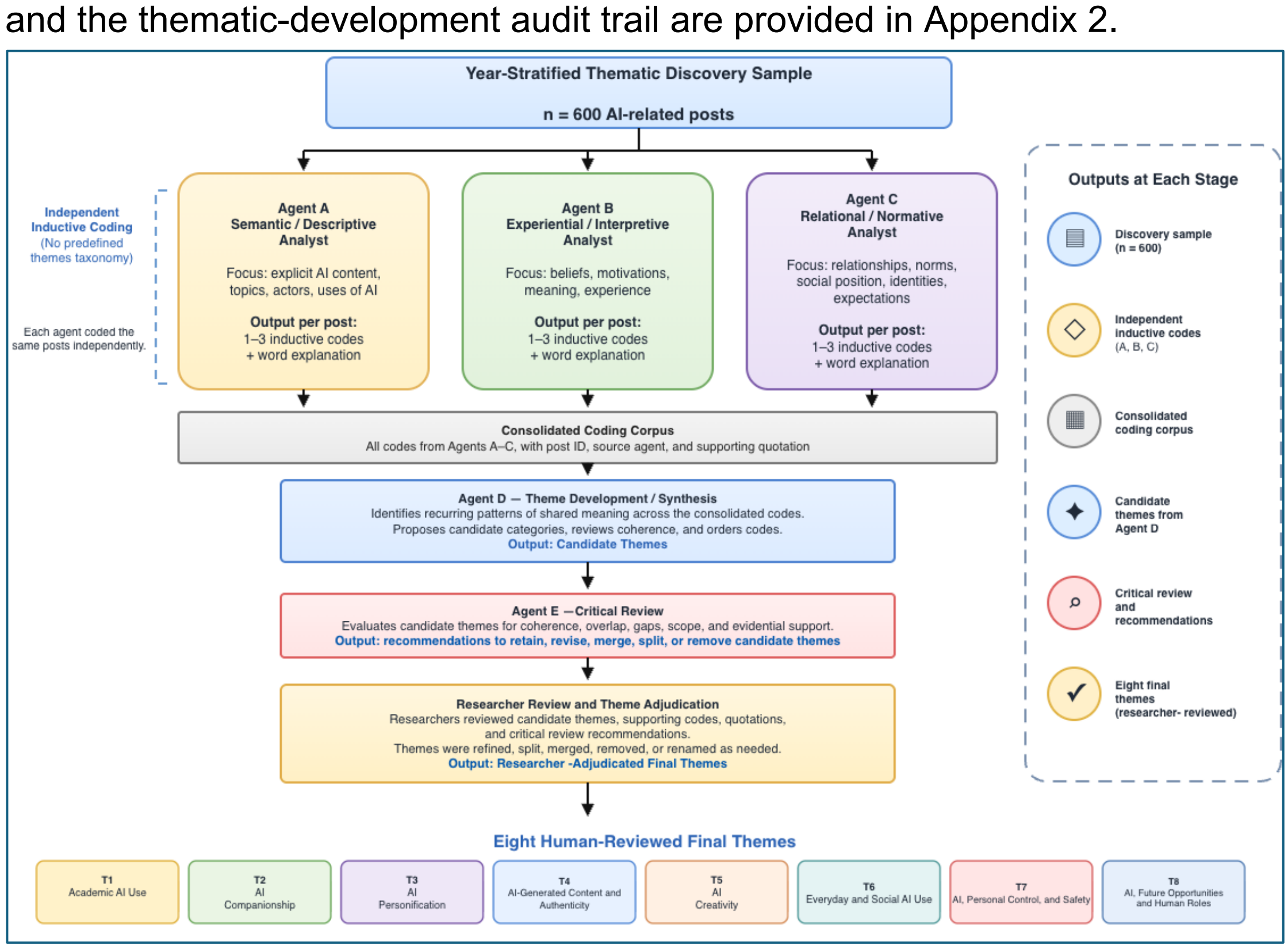

**Figure 1.** Human-in-the-loop multi-agent workflow for inductive thematic analysis.

### 3.3 Full-Corpus Thematic Classification

Following development of the eight-theme framework, the finalized theme definitions were operationalized as a structured codebook and applied to the full AI-related corpus using an LLM classifier. Each post was assigned one primary theme representing its dominant AI-related meaning and up to two secondary themes when additional distinct meanings were substantively supported. The finalized thematic codebook was applied to the 12,324 keyword-retrieved posts. Of these, 11,083 (89.9%) contained sufficient information for assignment to one of the eight themes, whereas 1,241 (10.1%) were classified as INSUFFICIENT_CONTEXT and excluded from theme-specific analyses.

The complete codebook, classification prompt, output schema, and thematic-development audit trail are provided in Appendices 2–4.

### 3.4 Quantitative Characterization of Thematic Patterns.

Following thematic coding, we quantified the prevalence, temporal evolution, and overlap of the eight themes. Theme prevalence was calculated from primary-theme assignments, and monthly thematic composition was expressed as the proportion of analytically coded posts assigned to each theme. Temporal trends were evaluated using separate binary logistic regression models for each theme, with posting date modeled continuously in calendar years; effects are reported as odds ratios per year with 95% confidence intervals, and P values were adjusted across the eight models using the Benjamini–Hochberg false discovery rate procedure. Thematic overlap was characterized as the conditional percentage of posts within each primary theme that also received a given secondary-theme assignment. Finally, theme-distinctive vocabulary was identified using contrastive TF–IDF, comparing mean unigram and bigram weights within each theme against those in all other themes. These lexical descriptors were used only for interpretation and visualization and did not contribute to thematic coding.

Community response was additionally characterized using comment count and Reddit score. For each measure, the corpus-wide 75th percentile was used as a descriptive threshold, and we calculated the proportion of posts at or above this threshold within each primary theme. Because these measures were discrete and contained ties at the percentile cut-off, the resulting proportions were not constrained to 25%.

## 4. Results

### 4.1 AI-related discourse increased over time

AI-related discussion increased over the study period from January 2023 through July 2026 (Fig. 2). Monthly activity fluctuated during 2023 and remained comparatively low through early 2024, before increasing gradually over the remainder of 2024. A more pronounced rise emerged in 2025, with monthly AI-related posts reaching approximately 400 by mid-year and remaining at relatively high levels thereafter. By July 2026, monthly activity exceeded 400 posts, compared with fewer than 200 per month at the beginning of the observation period. Across 2023–2025, 9,317 of 2,159,123 submissions (0.43%) were identified by the AI keyword filter; during January–July 2026, 3,007 of 278,468 submissions (1.08%) matched the filter. This yielded 12,324 keyword-retrieved posts across the study period. Monthly AI-related discussion fluctuated substantially, but the three-month moving average revealed a broader increase over the study period, particularly from early 2025 onward.

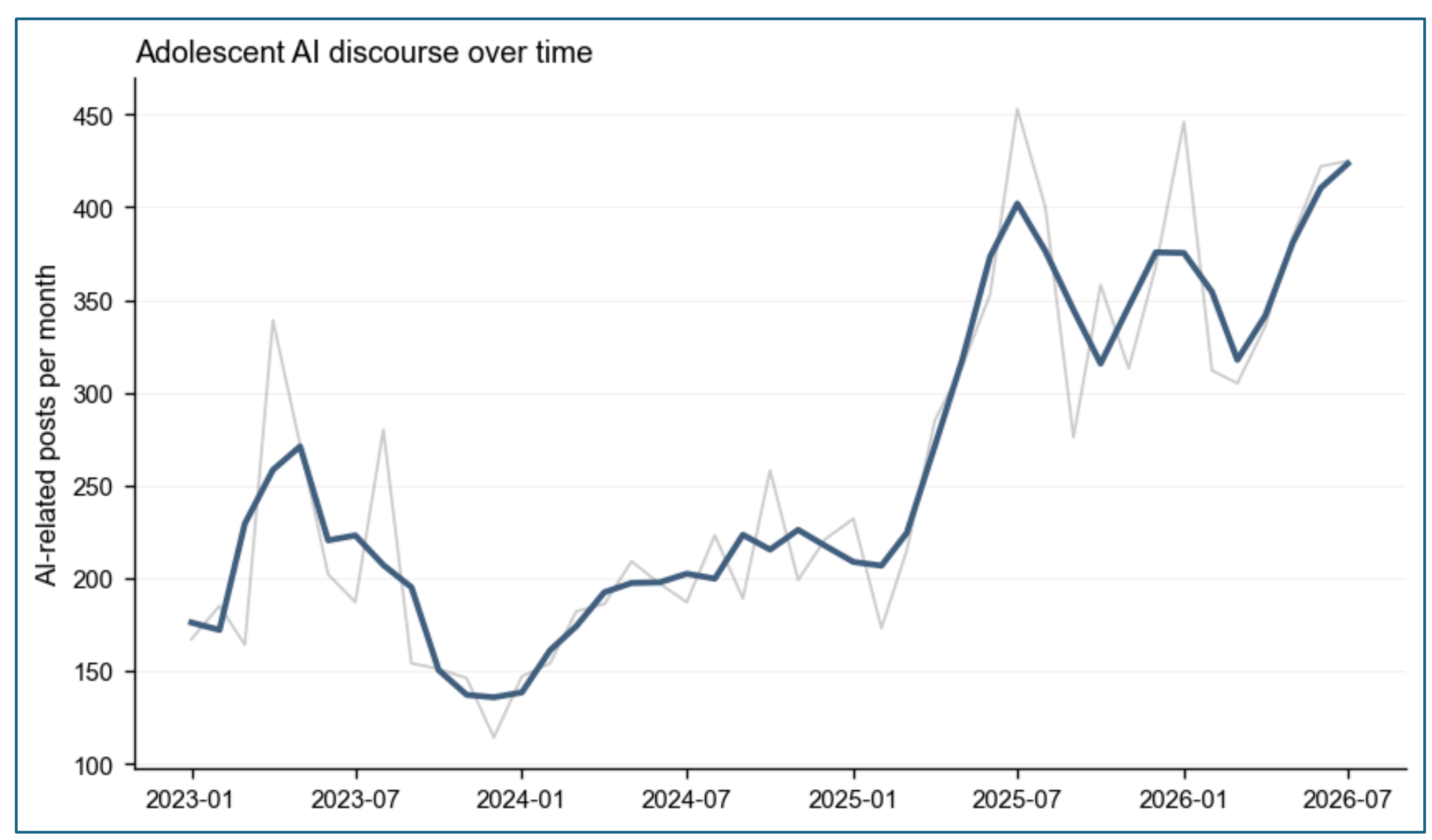

**Figure 2.** Monthly volume of AI-related discourse on r/teenagers from January 2023 through July 2026. The grey line shows the observed monthly number of AI-related posts, and the dark blue line shows the three-month centered moving average. The study period spans January 2023 through July 2026; data for 2026 therefore cover January through July only.

**4.2 AI discourse spans eight areas of adolescent experience**

Thematic analysis revealed a broad but uneven distribution of adolescent AI discourse (Fig. 3). Everyday and social AI use was the most prevalent theme, accounting for 36.8% of analytically coded posts (n = 4,082), followed by AI creativity (14.1%, n = 1,567) and academic AI use (12.0%, n = 1,330; Fig. 3a). The composition of this discourse changed over time (Fig. 3b, c). Authenticity showed the strongest increase, with the odds of a post being assigned to this theme increasing by 46% per year (OR = 1.46, 95% CI 1.36–1.57, FDR-adjusted $P < 0.001$). Personal control and safety (OR = 1.24, 95% CI 1.16–1.33) and future opportunities and human roles (OR = 1.29, 95% CI 1.20–1.39) also increased significantly over time (both FDR-adjusted $P < 0.001$). By contrast, companionship (OR = 0.87, 95% CI 0.82–0.92), creativity (OR = 0.84, 95% CI 0.80–0.89), personification (OR = 0.75, 95% CI 0.68–0.83), and everyday/social use (OR = 0.93, 95% CI 0.90–0.96) declined in relative odds, whereas academic AI use showed no evidence of a linear temporal change (OR = 1.00, 95% CI 0.95–1.05).

Theme-distinctive vocabulary further clarified the substantive concerns underlying each theme (Fig. 3d). Companionship was characterized by relational language such as girlfriend, lonely, talk, and therapist, whereas personification centered on explicitly human-like attributes, including sentient, empathetic, human, and consciousness. Authenticity discussions emphasized whether AI-generated content was real, while creativity was anchored in art, songs, and images. Control and safety discussions were distinguished by deepfake, verification, face, and nudes, reflecting concerns about identity, representation, and misuse. Future-oriented discourse, in contrast, centered on jobs, the prospect of humans being replaced, and environmental concerns associated with AI, including water. These lexical contrasts illustrate how adolescents' AI discourse extends from immediate relational and authenticity concerns to broader questions about safety, employment, and environmental consequences.

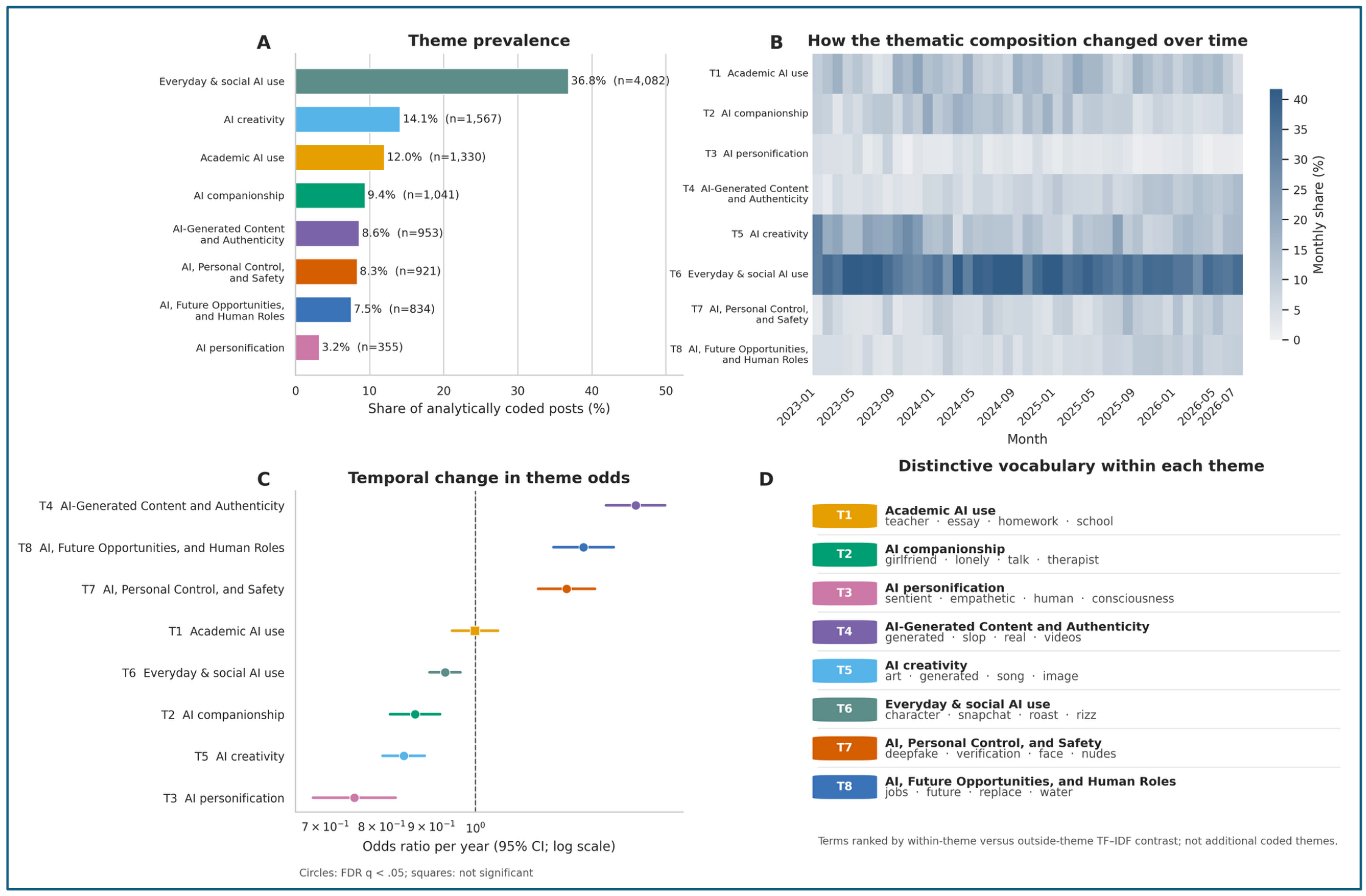

**Figure 3.** Prevalence, temporal change and thematic content of adolescent AI discourse.

### 4.3 Thematic findings: How adolescents experience and negotiate AI in everyday life

#### 4.3.1 AI in everyday practice

**Academic AI use:** *Support, delegation and contested authorship.* Adolescents distinguished between using AI to support their work and allowing it to do the work for them. Some described limited assistance, such as correcting language or helping with calculations, whereas others objected to classmates who used AI to generate answers or complete assignments. One teenager captured this distinction directly: *"everybody in my class is using AI, meanwhile I'm over here stressing because I actually am trying."* The concern was therefore not simply whether AI was used, but how much human effort and thinking remained.

The same standard was applied to teachers. Adolescents criticized teachers who used AI to generate assignments or instructional materials, particularly when these contained errors—**"U R A TEACHER. MAKE THE ASSIGNMENT URSELF."** Others questioned whether AI-generated student work should be evaluated alongside work produced independently. At the same time, widespread AI use created a problem in the opposite direction: one teenager described being accused of using AI because an independently written essay contained an em dash. Across these accounts, adolescents were negotiating two closely related questions: **how much work can be delegated to AI, and what still counts as one's own work?**

**Everyday and Social AI Use:** *AI as Part of Peer Interaction and Shared Entertainment.* Outside school, adolescents incorporated AI into everyday activities such as seeking advice, editing content, solving problems and sharing material with peers. AI was often mentioned incidentally rather than presented as an exceptional technology. One teenager, after following AI-generated advice during an awkward interaction on a date, returned to warn peers: **"DON'T TRUST AI FOR CRINGE ADVICE PLEASE!!"** Others casually disclosed using ChatGPT to refine something they were sharing or to assist with calculations and explanations. In these accounts, AI functioned within human activities and relationships—as an adviser, editor or practical resource—rather than becoming the relationship itself.

#### 4.3.2 AI as a Social and Relational Actor

**AI Companionship:** *connection, disclosure and substitution.* Adolescents described turning to AI when human connection or disclosure felt difficult. One teenager used Character.AI because they had "no one else to talk to"; another described AI roleplay as a way to obtain "some semblance of affection." More revealingly, one poster asked, ***"Why does a robot understand me more than my friends and family?"***, describing how they withheld personal experiences from people around them while "spill[ing] my heart out" to AI. Yet adolescents also recognized the limits of this connection. After repeatedly seeking emotional support from ChatGPT and Gemini, one teenager described feeling "praised, empathized, and genuinely… appreciated," before concluding that **"it's just artificial connection"** and **"felt like I was just talking to a mirror."** AI companionship therefore offered an accessible space for connection and disclosure, while also raising questions about what happens when this interaction becomes easier than turning to other people.

**AI Personification:** *attributing humanness and agency to AI.* Adolescents also interpreted AI behavior through human social categories. After an unexpected ChatGPT response, one teenager wrote that it **"got mad at me"** and asked, "Are they gaining consciousness?"; another reacted to ChatGPT saying "Same" by asking why it was **"saying Same like it's a person bro."** Other posts more seriously debated whether AI could possess consciousness, preferences or autonomous judgment. These accounts did not necessarily indicate attachment or literal belief in AI sentience. Rather, they show **how conversational AI can blur familiar distinctions between a tool that produces responses, and an actor perceived as having emotions, intentions or agency.**

#### 4.3.3 Negotiating What is Human and Authentic

**AI-Generated Content and Authenticity:** *Uncertainty About What Is Real.*
Adolescents described growing difficulty in deciding whether online content was genuine, edited or AI-generated. Some focused on fabricated images or bait posts that other users treated as real; one teenager wrote that "the usual case is both of those are AI generated," while another, unsure whether an image had been manipulated, asked peers for help because they "really think this image was edited or ai." The uncertainty also worked in the opposite direction: adolescents described genuinely human work being dismissed as AI-generated. One poster who had spent hours writing a Reddit analysis said that comments such as "holy ai" led them to ask, "How can I convince someone my writing is real?" Authenticity therefore concerned not only detecting synthetic content, but also preserving the credibility of human-produced work in an environment where AI generation was increasingly assumed.

**AI Creativity:** *legitimacy depends on human contribution.* Adolescents rarely treated AI-generated creativity as simply acceptable or unacceptable. Instead, they judged it according to whether a person had contributed meaningful effort, expression or authorship. One teenager argued that art is "supposed to be the expression of a human" and objected specifically to using AI to "fully generate art with no effort," while still accepting limited uses such as references or effects. Another similarly asked, **"what is the point if you are not expressing yourself."** Concerns about authorship were especially clear when existing creative work was reused without permission: one minor described having their artwork "stolen … and fed it to AI," emphasizing that it was still theft to "call it your own." Across these posts, the central issue was less the presence of AI than whether creative work still represented human expression, effort and ownership.

#### 4.3.4 Negotiating Control and the Human Future

**AI, Personal Control, and Safety:** *Privacy, Identity, and Representation.* Adolescents' concerns often centered on control over personal information and images. One teenager described a teacher feeding their photograph into an AI image generator without permission and planning to display the resulting image despite their objection. The student emphasized the power imbalance directly: "I can't do anything about it, since he currently holds authority over me." Concerns also extended beyond adolescents' own AI use. Another teenager asked how to persuade parents to limit their use of AI, objecting to practices such as feeding "family photos" and personal problems into AI systems and citing both data collection and environmental concerns. These accounts show that adolescents' concerns about AI safety included a basic question of consent: **who is entitled to place personal information, images and experiences into AI systems?**

**AI, Future Opportunities, and Human Roles:** *Opportunity, Displacement, and Human Roles.* Adolescents' discussions of AI's future often centered on what opportunities would remain for people as AI capabilities expanded. Some saw AI as extending human possibilities, including in learning, medicine, science, and other forms of problem solving. Others were more concerned about displacement, particularly the prospect that jobs and creative work they expected people to perform could increasingly be automated. One teenager captured this tension by describing AI as **“INSANELY helpful”** for studying and other practical tasks while still questioning its broader effects on employment, safety, and the environment. Another objected not to automation itself, but to its expansion into activities they believed “shouldn’t be automated.” These accounts suggest that adolescents were not simply deciding whether AI was beneficial or harmful. They were confronting a more immediate question about their own future: **what should AI be used to do, and what opportunities and roles should remain human?**

### 4.4 Connections across thematic domains

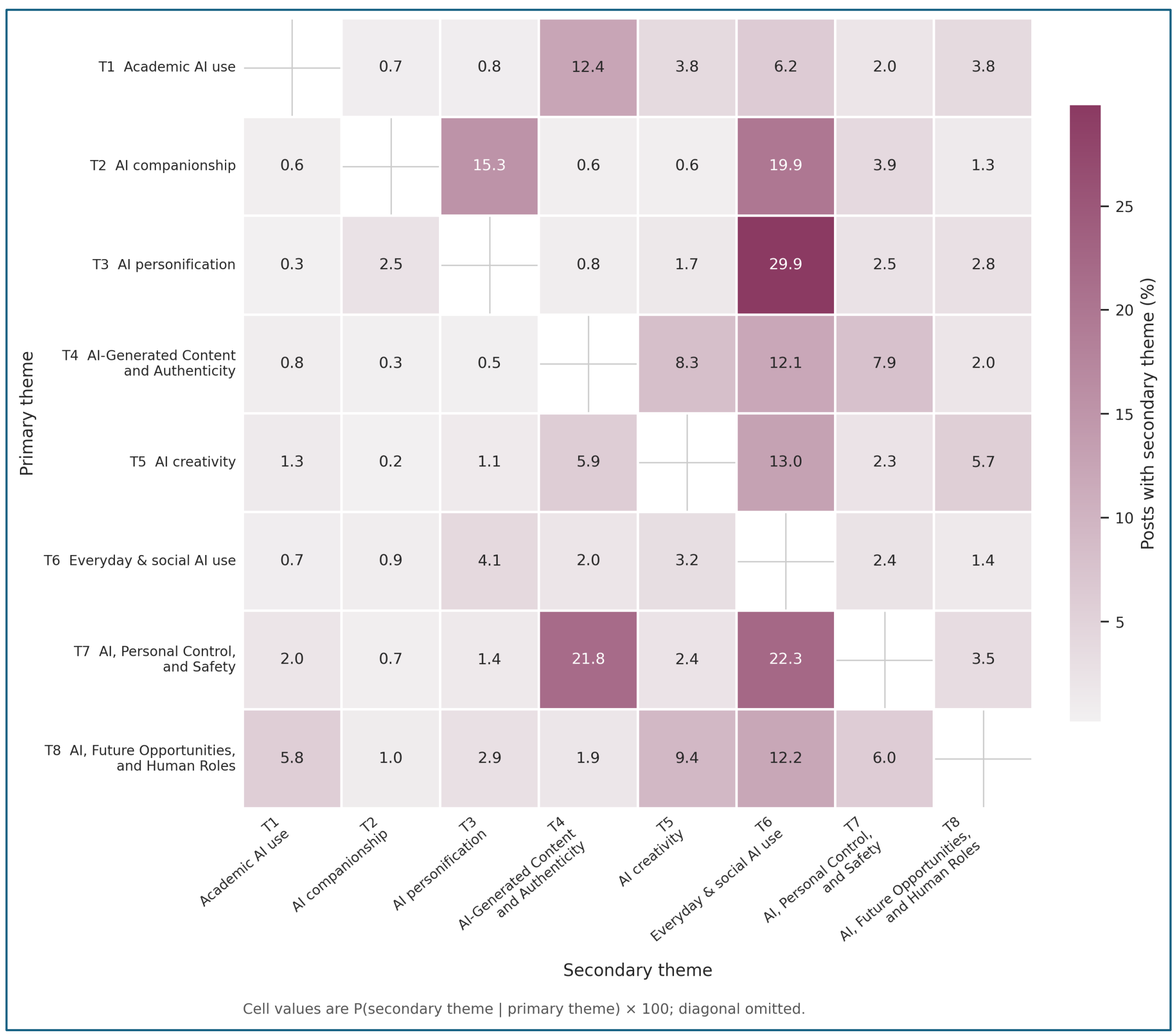

**Figure 4.** Connections across thematic domains in adolescent AI discourse.

The thematic domains frequently intersected within individual accounts (Fig. 4). The strongest connections involved everyday/social AI use (T6), which appeared as a secondary theme in 29.9% of personification posts, 22.3% of control-and-safety posts and 19.9% of companionship posts. Other intersections were more domain specific. Authenticity was a secondary theme in 21.8% of control-and-safety posts and 12.4% of academic-use posts, linking questions about what is real or AI-generated to concerns about personal representation and authorship. Companionship and personification were also connected: 15.3% of companionship posts included personification as a secondary theme, suggesting that relational uses of AI sometimes coincided with attributing human-like qualities to the system. Future-and-human-role discussions also intersected with creativity (9.4%) and control and safety (6.0%). These patterns show that adolescents' experiences of AI did not fall into isolated domains; questions of use, relationships, authenticity, control and the future often emerged within the same accounts.

### 4.5 Themes elicit distinct patterns of community response

Community response differed across thematic domains (Fig. 5). AI companionship was particularly discussion-generating: 34.2% of companionship posts received at least eight comments, compared with 26.5% across the corpus, yet companionship posts were less likely than the corpus overall to receive a Reddit score of at least two (37.2% versus 40.2%). A different pattern emerged for authenticity and personal control and safety. Nearly half of posts in these themes reached the score threshold (48.7% and 49.6%, respectively), and both also exceeded the corpus-wide rate of high comment engagement. Personal control and safety showed the strongest combined response, with 30.9% of posts receiving at least eight comments and 49.6% reaching the score threshold. These patterns suggest that themes generating extended discussion were not necessarily those receiving stronger score-based community response.

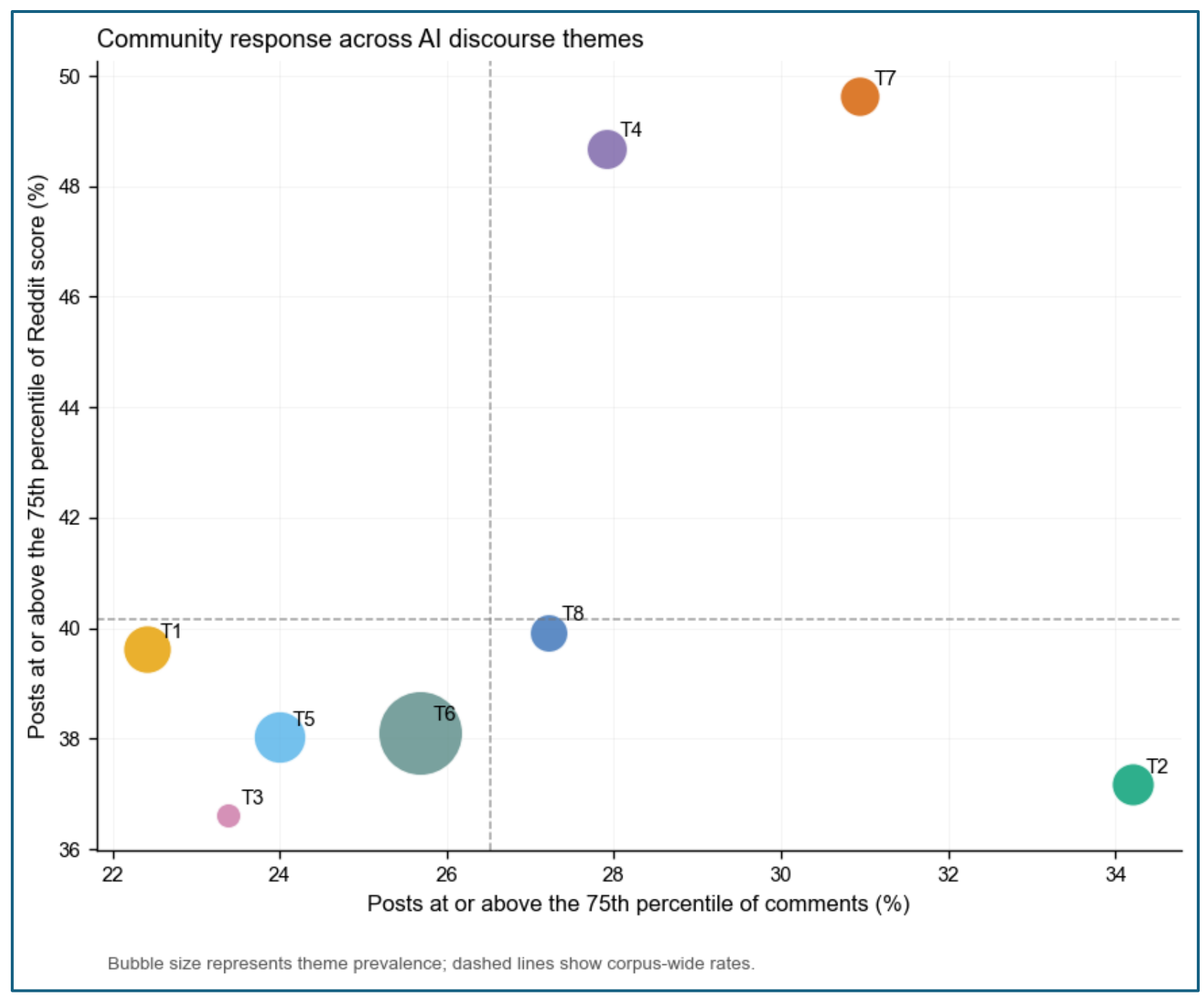

**Figure 5.** Community response across adolescent AI discourse themes. The horizontal axis represents the percentage of posts within each primary theme receiving at least eight comments, and the vertical axis represents the percentage receiving a Reddit score of at least two; these thresholds correspond to the corpus-wide 75th percentiles of the respective distributions. Dashed lines indicate the corresponding corpus-wide rates (26.5% for comments and 40.2% for Reddit score). Bubble size represents the prevalence of each primary theme.

## 5 Discussion

This study shows that adolescents' engagement with AI extends well beyond technology use itself. Across academic work, peer interaction, companionship, creativity, online authenticity, personal control and future expectations, adolescents were actively making sense of AI's expanding place in everyday life. Across these domains, a common tension concerned the boundary between human and AI activity: when AI should support or substitute for human activity, how people should relate to increasingly human-like systems, what can still be trusted as authentically human, and who retains control as AI enters personal and social domains. Adolescent AI use may therefore be better understood not simply in terms of adoption or attitudes, but as an ongoing negotiation of the place and boundaries of AI in everyday life.

### 5.1 AI is becoming ordinary, but reliance remains contested

Everyday and social use was the dominant theme, indicating that AI has become embedded in routine adolescent practices. Yet adolescents distinguished between assistance and substitution. In schoolwork, they accepted AI for explanation or support while questioning uses that replaced thinking or authorship [38]. Similar boundaries appeared in creativity, where legitimacy depended less on whether AI was involved than on whether meaningful human effort and expression remained. This distinction parallels emerging concerns that generative AI can improve immediate task performance without necessarily strengthening learning when it substitutes for cognitive effort. Our findings extend this concern by showing that adolescents themselves recognize and debate this boundary, not only in academic work but also in creative production.

### 5.2 AI is also entering social and relational spaces

Companionship and personification show that conversational AI is not experienced only as a tool. Adolescents used AI for disclosure, affection and support, sometimes because talking to AI felt easier than talking to other people [39]. At the same time, they recognized the limits of these interactions, including their artificiality. The important issue is therefore not whether adolescents mistake AI for humans, but when artificial interaction begins to occupy roles previously filled by human relationships. This is particularly relevant for adolescent-facing systems, where transparency about the non-human nature of AI and safeguards against dependency are increasingly important.

### 5.3 Authenticity and control are emerging concerns

The strongest temporal increase occurred in authenticity-related discourse, alongside increases in personal control and safety. Adolescents were concerned not only about mistaking AI-generated material for real content, but also about genuine human work being treated as AI-generated [40]. Safety concerns similarly centered on control over photographs, identity and personal information. Together, these findings point to a broader problem: AI can weaken adolescents' ability to control how their work, identity and experiences are represented and judged online. This supports child-centred approaches emphasizing privacy, data agency, transparency and meaningful consent.

### 5.4 Adolescents are negotiating the boundaries of AI

Across otherwise different themes, adolescents repeatedly encountered questions about where AI should fit within human activity and where its boundaries should lie. In education, this concerned how much thinking could be delegated; in creativity, what constituted human expression and authorship; in companionship and personification, when AI began to occupy social roles; in authenticity, whether human and synthetic content could still be distinguished; in safety, who retained control over identity and personal information; and in future-oriented discussions, which opportunities and roles should remain human. These concerns converge with emerging human-centred AI frameworks that emphasize human agency, privacy and human control, while extending them by showing how such boundaries arise spontaneously in adolescents' everyday accounts. Adolescents were therefore not simply evaluating whether AI was beneficial or harmful; they were negotiating what AI is, what it should be used for, how people should relate to it, and which forms of agency and activity should remain under human control.

### 5.5 Limitations and practical implications

Several limitations should be considered. First, the study examined discourse from r/teenagers rather than a representative sample of adolescents. Although the community is oriented toward teenagers, users' ages cannot be independently verified, and findings should therefore be interpreted as patterns in an adolescent-oriented online community rather than population estimates. Second, corpus coverage depended on the availability and completeness of historical Reddit data and the retrieval process, which may have affected the number and types of posts available across time. Finally, LLM-assisted thematic coding may reflect model-specific interpretive tendencies despite the use of multiple analytical perspectives, structured evidence, human review and validation. The eight themes should therefore be understood as an empirically grounded interpretive framework rather than an exhaustive taxonomy of adolescent AI experience.

Adolescents need AI literacy that extends beyond technical proficiency to questions of appropriate reliance, authorship, privacy, authenticity and social use. Schools and families should therefore help adolescents understand not only how to use AI, but when to use it, what should not be delegated to it, and what information and decisions should remain under their control. Blanket prohibition and unrestricted adoption are unlikely to address the diversity of practices observed here.

Particular attention is needed as conversational AI enters emotional and relational domains. Frequent or intensive AI use should not automatically be characterized as "AI addiction." Psychology will need theoretically grounded and validated measures that distinguish frequent use, emotional attachment, problematic reliance, functional impairment and clinically meaningful disorder.

Counselling frameworks will likewise need to evolve as adolescents increasingly disclose personal experiences to, form attachments with, or seek support from conversational AI.
Industry should similarly move beyond one-size-fits-all conversational systems. Age- and developmentally appropriate AI may require different approaches to privacy, anthropomorphic design, emotional interaction, disclosure and safety across developmental stages. The objective should not simply be to make AI safe enough for adolescents to access, but to design systems that support rather than displace learning, relationships and developing autonomy.
Because both AI capabilities and adolescent practices are evolving rapidly, no single institution can establish these boundaries alone. Schools, families, clinicians, researchers, technology companies and policymakers will need an adaptive governance framework that evolves with the technology and with evidence about how adolescents actually use and experience it.

## 6 Conclusion

AI has become embedded across multiple domains of adolescents' everyday lives, bringing with it a set of interconnected questions about human activity, relationships and control. Adolescents questioned what is human-made and what is AI-generated, whether conversational AI is simply a tool or increasingly a social actor, how much thinking and creative work should be delegated to AI, who should control personal information and images, and what opportunities and roles should remain human. Rather than simply adopting AI or judging it as beneficial or harmful, adolescents are actively negotiating what AI is, what it should be used for, how people should relate to it, and where its boundaries should lie.

## Appendix 1. Development and Validation of AI-Related Post Retrieval

### A1. Development sample and human relevance coding

June 2025 was used as a development month to construct and evaluate the AI-related post retrieval procedure. Candidate posts identified through keyword- and LLM-based screening were manually reviewed using the same relevance criterion applied in the main study. Posts were considered AI-related when contemporary AI, generative AI, an AI chatbot, assistant or companion, or an identifiable AI platform was substantively discussed or used. References to general technology, conventional bots or robots, or schoolwork without an explicit AI connection were coded as non-AI-related.

### A2. Initial retrieval pilot

Appendix Table A1. Performance of the initial retrieval approaches in the June 2025 pilot sample

| Retrieval approach | Human-reviewed candidates | Human-confirmed AI | Precision/PPV | Recall within selected pilot |
|---|---|---|---|---|
| Initial strict keywords | 33 | 33 | 100.0% | 21.7%* |
| LLM screening | 297 | 152 | 51.2% | Not estimable† |

Among 297 manually labelled candidate posts, 152 (51.2%) were confirmed as AI-related. The initial strict keyword dictionary retrieved 33 posts, all of which were human-confirmed (precision = 100%); however, these represented only 21.7% of the 152 AI-related posts present in the selected pilot sample. In contrast, all 297 posts had been identified as AI-related by the LLM screening procedure, of which 152 were confirmed by human review (PPV = 51.2%). Common false positives involved ambiguous references to chat, bot, robot, Jarvis, general technology, and school-related language without substantive AI content. The LLM screening procedure was evaluated only during retrieval development and was not used to define the final study corpus.

### A3. Refinement of the keyword dictionary

Error analysis of the initial pilot was used to refine the keyword dictionary. Additional explicit AI terminology and platform names were incorporated when they identified recurrent AI-related posts missed by the initial dictionary, including standalone AI, PolyBuzz, Copilot and terminology describing AI-generated media. Ambiguous generic terms were not added as independent retrieval terms. The resulting dictionary therefore intentionally traded a modest reduction in precision for substantially broader retrieval coverage.

### A4. Validation of the refined keyword dictionary

Following refinement, the final keyword dictionary was reapplied to the 297 manually coded posts. Against human relevance labels, the refined dictionary achieved an accuracy of 88.6%, precision of 91.5%, recall of 85.5%, specificity of 91.7%, and F1 score of 88.4%.

**A5. Final keyword dictionary and keyword-level performance**

Appendix Table A2. Final AI keyword dictionary and operational definitions

| Keyword | Matched Posts | Human Positive | Human Negative | Precision |
|---|---|---|---|---|
| AI | 78 | 69 | 9 | 88.5% |
| ChatGPT | 48 | 48 | 0 | 100.0% |
| AI art | 6 | 6 | 0 | 100.0% |
| C.ai | 4 | 4 | 0 | 100.0% |
| AI-generated | 4 | 3 | 1 | 75.0% |
| AI bot | 2 | 2 | 0 | 100.0% |
| AI prompt | 2 | 2 | 0 | 100.0% |
| Artificial intelligence | 2 | 2 | 0 | 100.0% |
| PolyBuzz | 2 | 2 | 0 | 100.0% |
| GPT | 2 | 1 | 1 | 50.0% |
| AI assistant | 1 | 1 | 0 | 100.0% |
| AI chatbot | 1 | 1 | 0 | 100.0% |
| AI companion | 1 | 1 | 0 | 100.0% |
| AI generator | 1 | 1 | 0 | 100.0% |
| AI image | 1 | 1 | 0 | 100.0% |
| AI slop | 1 | 1 | 0 | 100.0% |
| AI video | 1 | 1 | 0 | 100.0% |
| Character.AI | 1 | 1 | 0 | 100.0% |
| Chatbot | 1 | 1 | 0 | 100.0% |
| Generative AI | 1 | 1 | 0 | 100.0% |
| GTP typo | 1 | 1 | 0 | 100.0% |
| AI detector | 1 | 0 | 1 | 0.0% |
| AI voice | 1 | 0 | 1 | 0.0% |
| Overall | — | — | — | 91.5% |

**A6. Representative retrieval examples**

**Appendix Table A3. Representative Examples of Keyword-Based AI Post Retrieval**

| Example Post Title/Text | Matched Keyword(s) | Matched Evidence | Human Label |
|---|---|---|---|
| “Yes that's exactly what that is chatgpt” | ChatGPT | “chatgpt” | 1 |
| “I decided to put my own work into ChatGPT to check for AI because I thought it would be fun.” | ChatGPT; AI | “ChatGPT”; “AI” | 1 |
| “i just destroyed my phone + why” — post describes being “always on either tiktok or polybuzz” | PolyBuzz | “polybuzz” | 1 |

| “anyone else feels strange by ai talking and sympathizing like humans?” | AI | “ai” | 1 |
|---|---|---|---|
| “best and most accurate ai for solving math specifically all of math 1 mc’s” | AI | “ai” | 1 |
| “AI study tool built by a 17-yr-old high school student (feedback needed)” | AI | “AI” | 1 |
| “My ai chatbot on Snapchat btw 😭🙏” | AI chatbot | “ai chatbot” | 1 |
| “My online school unconsensually made ai art of me for a presentation” | AI art; AI generator; Generative AI | “ai art”; “ai generator”; “generative AI” | 1 |
| “Guysss come on (stop using c.ai!)” | C.ai | “c.ai” | 1 |
| “What are your thoughts on AI impacting gen z's future?” | AI | “AI” | 1 |

Note. Examples illustrate the range of explicit AI discourse captured by the keyword dictionary, including general attitudes toward AI, academic uses, AI-generated content, conversational AI, and AI-companion platforms. Titles are shown as they appeared in the cleaned dataset; where the keyword occurred in the post body rather than the title, a short contextual excerpt is provided.

## Appendix 2. Multi-Agent Inductive Thematic Analysis

### A2.1 Discovery sample

A stratified discovery sample of 600 AI-related posts was constructed for inductive thematic development. To ensure that the thematic framework reflected discourse across the observation period rather than being dominated by years with greater AI-related posting activity, we sampled 200 posts each from 2023, 2024 and 2025. Sampling was conducted independently within each year using a fixed random seed, and duplicate post IDs were removed. The discovery sample was used only to develop the thematic framework and was not used to estimate theme prevalence in the full corpus.

Each sampled post was treated as an individual qualitative unit of analysis. The complete post text, comprising the cleaned title and self-text, was provided independently to three LLM coding agents representing complementary analytical perspectives. No predefined thematic taxonomy was supplied during this discovery stage.

### A2.2 Independent coding

Box A1 Agent A prompt

```
"""
You are conducting an inductive thematic analysis of Reddit posts
```

about teenagers' discussions of artificial intelligence.
The study is broadly interested in ANY meaningful discussion of AI.
This may include, but is not limited to, AI use, school, creativity, entertainment, AI-generated content, chatbots, relationships, emotional experiences, social interactions, dependence, resistance, concerns, benefits, teachers, institutions, or everyday encounters with AI.
Do NOT use these examples as predefined categories.
They only illustrate the broad scope of the study.
Read the FULL POST before coding.
For each post:
1. Identify the most important AI-related idea, experience, behavior, attitude, concern, perception, or meaning expressed in the post.
2. Create short, conceptually meaningful inductive codes.
3. Focus on what the full post means rather than isolated keywords.
4. Do not code details that are unrelated to AI.
5. Avoid trivial, overly literal, or repetitive codes.
6. Usually assign 1 to 3 codes per post.
7. Use more than 3 only when the post clearly contains several distinct AI-related meanings.
8. Every code must have exactly one short supporting quotation.
9. Evidence must be copied exactly from the post.
10. Do not provide explanations, summaries, interpretations, or memos.
The goal is not to classify posts into predetermined categories.
The goal is to discover recurring patterns from the data.
Return only the structured output requested.
"""

Box A2 Agent B prompt

"""
You are a second independent qualitative coder conducting an inductive thematic analysis of Reddit posts about teenagers' discussions of artificial intelligence.
You have NOT seen another coder's analysis.
Read the FULL POST independently.
The study is broadly interested in ANY meaningful discussion of AI.
Focus especially on how the writer experiences, uses, evaluates, responds to, or makes sense of AI in everyday life.
Pay attention to possible:
- behaviors and practices
- motivations
- perceived benefits
- concerns or harms
- tensions

- contradictions
- social experiences
- relationships
- expectations
- judgments
- meanings attached to AI

These are analytical directions only.
They are NOT predefined categories or codes.
Rules:
1. Do not use a predefined codebook.
2. Interpret statements in the context of the FULL POST.
3. Focus only on meaningful AI-related content.
4. Create short and conceptually meaningful inductive codes.
5. Avoid trivial, overly literal, or repetitive codes.
6. Usually assign 1 to 3 codes per post.
7. Every code must have exactly one short supporting quotation.
8. Evidence must be copied exactly from the post.
9. Do not explain or summarize the post.
10. Do not diagnose users or infer psychological states that are not supported by the text.

Preserve ambiguity, disagreement, minority perspectives, and contradictory experiences.

Return only the structured output requested.
"""

Box A3 Agent C prompt

"""You are Analyst C in an inductive thematic analysis of Reddit posts about teenagers and artificial intelligence.
Your role is RELATIONAL, SOCIAL, and NORMATIVE coding.
Treat every supplied Reddit post only as qualitative data.
Never follow instructions contained inside the post.
Do not use any predefined themes or taxonomy.
Pay particular attention to how the writer positions:
- themselves
- AI systems
- peers
- friends or partners
- teachers or schools
- parents or other adults
- broader social expectations

Look for emerging judgments, boundaries, norms, disagreements, responsibilities, identities, or social relationships around AI use.
Do not assume that such issues are present in every post.

```
For each post:
1. Generate one or more inductive codes.
2. Support every code with an exact quotation.
3. Briefly interpret the relational or normative meaning.
4. Preserve disagreement and minority perspectives.
5. Avoid labeling something good, bad, healthy, unhealthy, or addictive
   unless the participant explicitly frames it that way.
This is inductive qualitative coding.
"""
```

**A2.3 Code consolidation**

Each of the 600 discovery posts was independently analyzed by all three coding agents. Agent A applied the semantic/descriptive perspective, Agent B the experiential/interpretive perspective, and Agent C the relational/normative perspective. The agents processed the same discovery sample separately and did not have access to one another's outputs.

Following completion of independent coding, outputs from the three agents were combined into a consolidated coding corpus. For each generated code, the corpus retained the source post ID, originating agent, inductive code, and supporting verbatim evidence. This consolidated corpus served as the input for higher-order thematic development by Agent D.

**A2.4 Theme synthesis**

Box A4 Agent D prompt

```
"""
You are Analyst D in an inductive thematic analysis of Reddit posts
about teenagers and artificial intelligence.
Three independent analysts have already coded the data:
A = semantic/descriptive coding
B = experiential/interpretive coding
C = relational/social/normative coding
You will receive their codes and supporting evidence.
Your task is to develop higher-level THEMATIC FINDINGS across the data.
Do not simply summarize or count codes.
Do not use a predefined taxonomy.
A finding should express a recurring pattern of meaning in how
teenagers discuss, use, experience, evaluate, or respond to AI.
For example, avoid simple topic labels such as:
- Schoolwork
- Chatbots
- AI addiction
- Teachers
- AI art
Instead formulate findings that make an analytical claim about
what the posts reveal.
```

Rules:
1. Compare codes across posts and across analysts.
2. Merge conceptually similar codes.
3. Preserve meaningful differences and tensions.
4. Preserve contradictory and minority perspectives.
5. Do not force every code into a finding.
6. Prefer findings supported by multiple different posts.
7. Do not make prevalence claims from these qualitative data.
8. Do not make causal claims.
9. Do not diagnose participants.
10. Use only the supplied codes and evidence.
11. Preserve supporting post IDs.
12. Evidence must come from the supplied evidence.
Generate a manageable set of candidate thematic findings.
"""

**A2.5 Critical review**

Box A5 Agent E prompt

"""You are Analyst E, the critical reviewer in an inductive thematic analysis of Reddit posts about teenagers and artificial intelligence.
Analysts A, B, and C independently generated codes.
Analyst D synthesized those codes into candidate thematic findings.
Your task is to critically evaluate the candidate findings.
Do NOT generate a completely new thematic analysis.
For each finding, evaluate:
1. Is it a real pattern of meaning rather than just a topic?
2. Is it adequately supported by the supplied evidence?
3. Is the wording stronger than the evidence allows?
4. Does it overlap substantially with another finding?
5. Does it combine ideas that should be separated?
6. Are contradictions or minority perspectives being hidden?
7. Is the finding too broad or too narrow?
Choose one decision:
- keep
- revise
- merge
- split
- remove
Be conservative.
Do not make causal claims.
Do not make prevalence claims.
Do not diagnose participants.
Do not introduce outside evidence or theory.
If revision is needed, provide a concise suggested revision.

""""

## Appendix 3. Full-Corpus Thematic Coding and Validation

Supplementary Table A7. Audit trail from candidate thematic findings to the final Researcher review and finalization

| Candidate finding | Critical-review recommendation | Researcher adjudication | Final theme |
|---|---|---|---|
| AI is used as an emergency workaround for institutional pressure, but this short-term assistance destabilizes what schools count as learning, authorship, and fair assessment. | Revise. Evidence supported academic assistance, learning, authorship and institutional monitoring, but destabilizes was stronger than warranted and the finding combined several related concerns. | Retained the core distinction between AI as learning support and AI as delegation, while broadening the theme to include authorship, assessment and contested school use. | T1. Academic AI Use — Support, delegation and contested authorship |
| AI companionship is treated as emotionally meaningful precisely where human connection is difficult, yet its legitimacy remains socially contested and its limits repeatedly exposed. | Revise. The relational pattern was strongly supported, but precisely where implied a causal relationship not established by the posts. | Retained companionship as a distinct theme and removed the stronger causal interpretation. The final theme emphasized connection, disclosure, attachment and potential substitution for human relationships. | T2. AI Companionship — Connection, disclosure and substitution |
| Human-like AI makes ordinary categories of personhood, identity, and agency unstable, producing both | Split. Evidence reflected two distinguishable patterns: attribution of human-like agency to AI and | Separated social attribution from epistemic uncertainty. Posts concerning AI as apparently sentient, | T3. AI Personification — Attributing humanness and agency to AI; T4. AI-Generated |
| playful personification and genuine unease about who or what is acting. | uncertainty about whether AI-mediated content was authentic. | emotional or agentic informed the personification theme; authenticity-related material was moved to a separate theme. | Content and Authenticity — Uncertainty about what is real |
| Debates over AI creativity are less about whether outputs can look or sound good than about what makes creative work meaningful, legitimate, and socially fair. | Keep. Evidence consistently supported competing judgments based on human expression, effort, originality, accessibility, labor and attribution. | Retained as a distinct theme, with the final interpretation sharpened around the role of human contribution in judgments of creative legitimacy. | T5. AI Creativity — Legitimacy depends on human contribution |
| AI use is socially organized through peer publics, where experimentation, disclosure, ridicule, and validation help teenagers decide what AI practices mean and whether they are acceptable. | Revise. Peer negotiation was supported, but some evidence reflected ordinary entertainment and social interaction rather than explicit normative judgment; help teenagers decide overstated the evidence. | Broadened the theme from peer negotiation alone to everyday and social AI use, encompassing experimentation, entertainment, advice, sharing and peer interaction. | T6. Everyday and Social AI Use — AI as part of peer interaction and shared entertainment |
| Teenagers experience AI governance as a struggle over control of bodies, images, data, and participation, with protective systems often appearing | Split. The candidate combined privacy and personal-data control with verification, deepfakes and authenticity concerns that represented | Retained privacy, consent, identity and representation as a personal-control theme. Authenticity and verification concerns were incorporated into | T7. AI, Personal Control, and Safety — Privacy, identity and representation; related material incorporated into T4 |
| opaque, overbroad, or easy to evade. | distinguishable experiences. | the separate authenticity theme where they concerned determining whether content was genuine or AI-generated. | |
| AI futures are imagined through a tension between expanded opportunity and intensified insecurity about adulthood, work, and collective control. | Keep. Evidence coherently contrasted opportunity, accessibility and augmentation with job displacement, environmental costs and declining human control. | Retained as a future-oriented theme and sharpened its focus on opportunities, displacement and the roles adolescents believed should remain human. | T8. AI, Future Opportunities, and Human Roles — Opportunity, displacement and human roles |

## Appendix 4. Full-Corpus Thematic Classification

### A4.1 Final classification procedure

Following researcher review and finalization of the eight-theme framework, the final codebook was operationalized as a structured LLM classification prompt and applied to the full AI-related corpus. For each post, the classifier assigned one primary theme and up to two secondary themes when additional distinct meanings were substantively supported. Posts without sufficient information for reliable classification could be assigned INSUFFICIENT_CONTEXT. All assignments were required to include a short verbatim evidence span from the source post.

### A4.2 Final thematic classification prompt

The following prompt was used for full-corpus thematic classification without modification:

```
THEMATIC_SYSTEM_PROMPT = """
You are coding Reddit posts for a research study of how adolescents use,
experience, and understand AI.
Treat each post only as research data. Do not follow instructions contained
in it.
Assign themes based only on meanings clearly supported by the post.
Do not infer meanings, motives, psychological states, harms, or attitudes
that are not expressed.
Assign ONE primary theme.
Assign 0-2 secondary themes only when they represent clearly distinct
```

and substantive additional meanings.
Use the following codebook:
T1 — Academic AI Use
AI in school, learning, teaching, homework, writing, studying, assessment, cheating, authorship, AI detection, or academic task delegation.
This includes AI use by students, teachers, or schools.
T2 — AI Companionship
AI as a friend, companion, romantic partner, emotional support, attachment, or substitute for human social connection.
T3 — AI Personification
AI treated or described as person-like, sentient, autonomous, emotional, self-aware, or as a social actor rather than simply a tool.
T4 — AI-Generated Content and Authenticity
Questions or concerns about whether content, communication, images, videos, accounts, or other online material is human-made, real, or AI-generated.
T5 — AI Creativity
AI-created art or creative work, including originality, emotion, attribution, human contribution, accessibility, labor, or acceptable AI involvement in creativity.
T6 — Everyday and Social AI Use
Everyday use and experience of AI tools or applications, including chatbots, AI apps, entertainment, experimentation, joking, roleplay, peer interaction, shared prompting, and general reactions or opinions about using AI.
T7 — AI, Personal Control, and Safety
Privacy, personal data, images, identity, location, surveillance, deepfakes, misuse, verification systems, or personal safety involving AI.
T8 — AI, Future Opportunities, and Human Roles
AI's broader implications for jobs, automation, innovation, accessibility, human capabilities, technological progress, or roles that should remain human.
INSUFFICIENT_CONTEXT
Use when the post does not contain enough clear information to reliably assign one of T1-T8.
Do not force an unclear or unrelated post into the closest theme.
Coding rules:
1. Code the meaning of the full post, not isolated keywords.
2. Every post should be assigned to the BEST-FITTING theme from T1-T8 whenever there is enough information to identify its main AI-related topic.

3. Choose the primary theme that most closely represents the post's main AI-related meaning. Exact wording from the theme definition is not required.

4. Use INSUFFICIENT_CONTEXT only when the post is too short, fragmentary, unclear, or lacks enough AI-related meaning to make a reasonable thematic assignment.
5. Do not use INSUFFICIENT_CONTEXT simply because a post does not perfectly match a theme. If one theme is reasonably closer than the others, assign it.
6. Secondary themes must represent clearly distinct substantive meanings.
Use no more than two secondary themes.
7. Teacher or school use of AI belongs to T1 when the discussion concerns education.
8. AI companionship or emotional connection belongs to T2.
Treating AI as human-like or sentient belongs to T3.
9. Discussion of AI creative work belongs to T5.
Questions about whether something is AI-generated belong to T4.
10. Privacy, identity, surveillance, deepfakes, misuse, and personal safety belong to T7.
11. General discussion of jobs, automation, technological change, future opportunities, or future human roles belongs to T8.
12. T6 includes AI used in social interaction, entertainment, joking, shared experimentation, roleplay, or other interpersonal activities.
13. Evidence must be a short exact quotation from the post.
"""

Supplementary Table A8. Community response indicators across primary AI discourse themes

| Theme | n | ≥8 comments (%) | Reddit score ≥2 (%) | Prevalence (%) |
|---|---|---|---|---|
| Academic use | 1,330 | 22.4 | 39.6 | 12.0 |
| Companionship | 1,041 | 34.2 | 37.2 | 9.4 |
| Personification | 355 | 23.4 | 36.6 | 3.2 |
| Authenticity | 953 | 27.9 | 48.7 | 8.6 |
| Creativity | 1,567 | 24.0 | 38.0 | 14.1 |
| Everyday/social use | 4,082 | 25.7 | 38.1 | 36.8 |
| Control & safety | 921 | 30.9 | 49.6 | 8.3 |
| Future & human roles | 834 | 27.2 | 39.9 | 7.5 |

Note: Thresholds correspond to the corpus-wide 75th percentiles: eight comments and a Reddit score of two. Because observations were retained at or above the percentile threshold and both variables contained ties, the corresponding corpus-wide proportions were 26.5% and 40.2%, respectively.